\documentclass[smallcondensed]{svjour3}

\usepackage{natbib}
\usepackage{graphicx}
\usepackage{algorithm}
\usepackage{algorithmic}
\usepackage{amssymb}

\usepackage{color}
\usepackage{soul}

\usepackage{amsmath}
\usepackage{bm}
\usepackage{dsfont}
\bibpunct{(}{)}{;}{a}{,}{,}

\newtheorem{defm}{Definition}

\title{Bayesian Policy Reuse}
\titlerunning{Bayesian Policy Reuse}

\author{
 Benjamin Rosman \and Majd Hawasly\footnote{The first two authors contributed equally to this paper.} \and Subramanian Ramamoorthy}
 \authorrunning{ Rosman, Hawasly \& Ramamoorthy }

 \institute{Benjamin Rosman \at Mobile Intelligent Autonomous Systems (MIAS), Council for Scientific and Industrial Research (CSIR), South Africa, and the School of Computer Science and Applied Mathematics, University of the Witwatersrand, South Africa.
\email{BRosman@csir.co.za}. \and Majd Hawasly \at School of Informatics, University of Edinburgh, UK.
\email{M.Hawasly@ed.ac.uk}. \and Subramanian Ramamoorthy \at School of Informatics, University of Edinburgh, UK.
\email{S.Ramamoorthy@ed.ac.uk}.}

\begin{document}

\maketitle

\begin{abstract}

A long-lived autonomous agent should be able to respond online to novel instances of tasks from a familiar domain. Acting online requires `fast' responses, in terms of 
rapid convergence, especially when the task instance has a short duration such as in applications involving interactions with humans. These requirements can be problematic for many established methods for learning to act. 
In domains where the agent knows that the task instance is drawn from a family of related tasks, albeit without access to the label of any given instance, it can choose to act through a process of \emph{policy reuse} from a library  in contrast to policy learning.
In policy reuse, the agent has prior experience from the class of tasks in the form of a 
library of policies that were learnt from sample task instances during an offline training phase. We formalise the problem of policy reuse  and present an algorithm for efficiently responding to a novel task instance by reusing a policy from this library of existing policies, where the choice is based on observed `signals' which correlate to policy performance. We achieve this by posing the problem as a Bayesian choice problem with a corresponding notion of an optimal response, but the computation of that response is in many cases intractable. Therefore, to reduce the computation cost of the posterior, we follow a Bayesian optimisation approach and define a set of policy selection functions, which balance exploration in the policy library against exploitation of previously tried policies, together with a model of expected performance of the policy library on their corresponding task instances.  We validate our method in several simulated domains of interactive, short-duration episodic tasks, showing rapid 
convergence in 
unknown task variations.


\end{abstract}

\section{Introduction}
As robots and software agents are becoming more ubiquitous in many applications involving human interactions, greater numbers of scenarios require new forms of decision making that allow fast responses to situations that may drift or change from their nominal descriptions.

For example, \emph{online personalisation} \citep{mahmud2014adapting} is becoming a core concept in human-computer interaction (HCI), driven largely by a proliferation of new sensors and input devices which allow for a more natural means of communicating with hardware. Consider, for example, an interactive interface in a public space like a museum that aims to  provide information or services to users through normal interaction means such as natural speech or body gestures. 
The difficulty in this setting is that the same device may be expected to interact with a wide and diverse pool of users, who differ both at the low level of interaction speeds and faculties, and at the higher level of which expressions or gestures seem appropriate for particular commands. 
The device should autonomously calibrate itself to the class of user, and a mismatch in that could result in a failed interaction. On the other hand, taking too long to calibrate is likely to frustrate the user~\citep{rosman2014user}, who may then abandon the interaction. 

This problem, characterised as a short-term interactive adaptation to a new situation (the user), also appears in interactive situations other than HCI. 
As an example, consider a system for localising and monitoring poachers in a large wildlife reserve\footnote{Poaching of large mammals such as rhinoceroses is a major problem throughout Africa and Asia \citep{amin2006rhino}.} that comprises an intelligent base station which can deploy light-weight airborne drones to scan particular locations in the reserve for unusual activity. 
While the tactics followed by the poachers in every trial would be different, the possible number of drone deployments in a single instance of this adversarial problem is limited, as the poachers can be expected to spend a limited time stalking their target before leaving. 

In this paper, we formalise and propose a solution to the general problem inspired by these real world examples. To this end, we present a number of simulated scenarios to investigate different facets of this problem, and contrast the proposed solution with related approaches from the literature.

The key component of this problem is the need for efficient decision making, in the sense that the agent is required to adapt or respond to scenarios which exist only for short durations.
  As a result, solution methods are required to have both \emph{low convergence time} and \emph{low regret}. To this end, the key intuition we employ is that nearly-optimal solutions computed within a small computational and time budget are preferred to those that are optimal but unbounded in time and computation. 
Building on this, the question we address in this paper is how to \emph{act well} (not necessarily optimally) in an \emph{efficient manner} (for short duration tasks) in a large 
space of qualitatively-similar tasks.

While it is unreasonable to expect that an arbitrary task instance  could be solved from scratch in a short duration task (where, in general, the interaction length is unknown), it is plausible to consider seeding the process with a set of policies of previously solved, related task instances, in what can be seen as a strategy for transfer learning~\citep{transfer}. In this sense, we prefer to quickly select a nearly-optimal pre-learnt policy, rather than learn an optimal one for each new task. For our previous examples, the interactive interface may ship with a set of different classes of user profiles which have been acquired offline, and the monitoring system may be equipped with a collection of pre-learnt behaviours to navigate the reserve when a warning is issued. 

We term this problem of short-lived sequential policy selection for a new instance 
the \emph{policy reuse} problem, which differs slightly from other uses of that term (see Section~\ref{sec:other_pr}), and we define it  as follows. 

\begin{defm}[Policy Reuse Problem] Let an agent be a decision making entity in a specific domain, equipped with a policy library $\Pi$ for some tasks in that domain. The agent is presented with an unknown task which must be solved within a limited, and small, number of trials. At the beginning of each trial episode, the agent can select one policy from $\Pi$ to execute for the full episode. The goal of the agent is thus to
\begin{quote}
{select policies for the new task from $\Pi$ to minimise the total regret, with respect to the performance of the best alternative from $\Pi$ in  hindsight, incurred in the limited task duration.}
\end{quote}
\label{def:pr}
\end{defm}

The online choice from a set of alternatives for minimal regret 
could be posed as a multi-armed bandit problem. Here, each arm corresponds to a pre-learnt policy
, and our problem becomes that of a sequential, finite-horizon, optimal selection from a fixed set of policies. 
Solving this problem in general is difficult as it maps into the intractable finite-horizon online bandit problem~\citep{finitebandits}. On the other hand, traditional  approaches to solve the multi-armed bandit problem 
involve testing each available arm on the new task in order to gauge its performance, which may be a very costly procedure from a 
convergence rate point of view. 

Instead, one can exploit knowledge which has been acquired offline to improve online response times. These more informed approaches to the multi-armed bandit problem exploit background domain information (e.g. contexts in contextual bandits~\citep{contextual,epoch_greedy}) or analytical forms of reward (e.g. correlated bandits~\citep{depndentbandits}) to share the credit of pulling an arm between many possible arms. This however requires prior knowledge of the domain and its metrics, how possible domain instances relate to the arms, and in some cases to be able to determine this side information for any new instance. 

We propose a solution for policy reuse that neither requires complete knowledge of the space of possible task instances nor a metric in that space, but rather builds a surrogate model of this space from offline-captured correlations between the policies when tested under canonical operating scenarios. Our solution then maintains a Bayesian belief over the nature of the new task instance in relation to the previously-solved ones. Then, executing a policy provides the agent with information which, when combined with the model, is not only useful to evaluate the policy chosen but also to gauge the suitability of other policies. This information  updates the belief, which facilitates choosing the next policy to execute. 

We treat the policy selection in the policy reuse problem as one of \emph{optimisation} of the response surface of the new task instance, although over a finite library of policies. Because we are dealing with tasks which we assume are of limited duration and which do not allow extensive experimenting, and in order to use information from previous trials to maintain belief distributions over the task space, we draw inspiration from the Bayesian optimisation/efficient global optimisation literature \citep{bayesopt} for an approach to this problem that is efficient in the number of policy executions, corresponding to function evaluations in the classical optimisation setting. 

\subsection{Other Definitions of Policy Reuse}
\label{sec:other_pr}

A version of the policy reuse problem was described by~\cite{jmlr}, where it is used to test a set of landmark policies retrieved through clustering in the space of MDPs. Additionally, the term `policy reuse' has been used by~\cite{policyreuse} in a different context. There, a learning agent is equipped with a library of previous policies to aid in exploration, as they enable the agent to  collect relevant information quickly to accelerate learning. In our case, we do not expect to have enough time to learn a full policy, and so instead rely on aggressive knowledge transfer using our proposed policy reuse framework to achieve the objective of the agent.



\subsection{Contributions and Paper Organisation}

The primary contributions made in this paper are as follows:

\begin{enumerate}
 \item We introduce Bayesian Policy Reuse (BPR) as a general Bayesian framework for solving the policy reuse problem as defined in Definition~\ref{def:pr}  (Section~\ref{sec:bpr}).
 \item We present several specific instantiations of BPR using different policy selection mechanisms (Section~\ref{sec:bprkg}), and compare them on an online personalisation domain (Section~\ref{sec:exp:personalisation}) as well as a domain modelling a surveillance problem (Section~\ref{sec:exp:surveillance}).
 \item We provide an empirical analysis of the components of our model, considering different classes of observation signal, and the trade-off between library size and convergence rate.
\end{enumerate}


\section{Bayesian Policy Reuse}
\label{sec:bpr}
We now pose the policy reuse transfer problem within a Bayesian framework. 
 Bayesian Policy Reuse (BPR) builds on the intuition that, in many cases, performance of a specific policy is better, relative to the other policies in the library, in tasks within some neighbourhood of the task for which it is known to be optimal
. Thus, a model that measures the similarity between a new task and other known tasks may provide indications as to which policies may be the best to reuse. We learn such a model from offline experience, and then use it online as a Bayesian prior over the task space, which is updated with new observations from the current task. Note that in this work we 
consider the general case where we do not have a parametrisation of the task space that allows constructing that model explicitly (e.g.~\cite{daSilva12a}). This may be the case where aspects of the model may vary qualitatively (e.g. different personality types), or where the agent has not been exposed to enough variations of the task to learn the underlying parametric model sufficiently.

\subsection{Notation}

Let the space of task instances be $\mathcal{X}$, and let a task instance $x\in\mathcal{X}$ be specified by a Markov Decision Process (MDP). An MDP is defined as a tuple $\mu = (S, A, T, R, \gamma)$, where $S$ is a finite set of states; $A$ is a finite set of actions which can be taken by the agent; $T : S \times A \times S \rightarrow [0,1]$ is the state transition function where $T(s,a,s')$ gives the probability of transitioning from state $s$ to state $s'$ after taking action $a$; $R : S \times A \times S \rightarrow \mathds{R}$ is the reward function, where $R(s,a,s')$ is the reward received by the agent when transitioning from state $s$ to $s'$ with action $a$; and finally, $\gamma \in [0,1]$ is a discounting factor. As $T$ is a probability function, $\sum_{s' \in S} T(s,a,s') = 1, \forall a \in A, \forall s \in S$. Denote the space of all MDPs $\mathcal{M}$. We will consider episodic tasks, i.e. tasks that have a bounded time horizon.

A policy $\pi : S \times A \rightarrow [0,1]$ for an MDP is a distribution over states and actions, defining the probability of taking any action from a state. The return, or utility, generated from running the policy $\pi$ in an episode of a task instance is the accumulated discounted reward, $U^\pi = \sum_{i=0}^{k} \gamma^i r_i$, with $k$ being the length of the episode and $r_i$ being the reward received at step $i$. We refer to $U^\pi$ generated from a policy $\pi$ in a task instance simply as the policy's \emph{performance}.
Solving an MDP $\mu$ is to acquire an optimal policy $\pi^* = \arg\max_\pi U^\pi$ which maximises the total expected return of $\mu$. For a reinforcement learning agent, $T$ and $R$ are typically unknown. We denote a collection of policies possessed by the agent by $\Pi$, and refer to it as the policy library. 


We complete the discussion of the formulation of a task with the definition of \emph{signals}. The aim of signals is to provide the agent with auxiliary information that hints toward identifying the nature of the new task instance in the context of the previously-solved instances.
\begin{defm}[Signal]
A \emph{signal} $\sigma \in \Sigma$ is any information which is correlated with the performance of a policy and which is provided to the agent in an online execution of the policy on a task.
\end{defm}
The most straightforward signal is the performance itself, unless this is not directly observable (e.g. in cases where the payoff may only be known after some time horizon
). The information content and richness of a signal determines how easily an agent can identify the type of the new task with respect to the previously-solved types. This is discussed in more detail in Section~\ref{sec:signals}.

Throughout the discussion, we adopt the following notational convention: $\mathrm{P}(\cdot)$ refers to a probability, $\mathrm{E}[\cdot]$ refers to an expectation, $\mathrm{H}(\cdot)$ refers to entropy, and $\Delta(\cdot)$ is a distribution.

\subsection{Overview}
For a set of previously-solved tasks $\mathcal{X}$ and a set of policies $\Pi$, Bayesian Policy Reuse involves two key probability models:
\begin{itemize}
 \item The first, $\mathrm{P}(U|\mathcal{X},\Pi)$, where $U\in\mathds{R}$ is utility, is the \emph{performance model}; a probability model over performance of the library of policies $\Pi$ on the set of previously-solved tasks $\mathcal{X}$. This information is available in an offline phase. 

\item The second key component is the \emph{observation model}, defined as a probability distribution $\mathrm{P}(\Sigma|\mathcal{X},\Pi)$ over $\Sigma$, the space of possible \emph{observation signals}. Any kind of information  that can be observed online and that is  correlated with the performance can be a used as an observation signal. When performance information is directly observable online (e.g., not delayed), performance can be used as the signal, and in this case the observation and the performance models can be the same. 

\end{itemize}

\begin{figure}[!h]
\centering
 \includegraphics[width=0.9\textwidth]{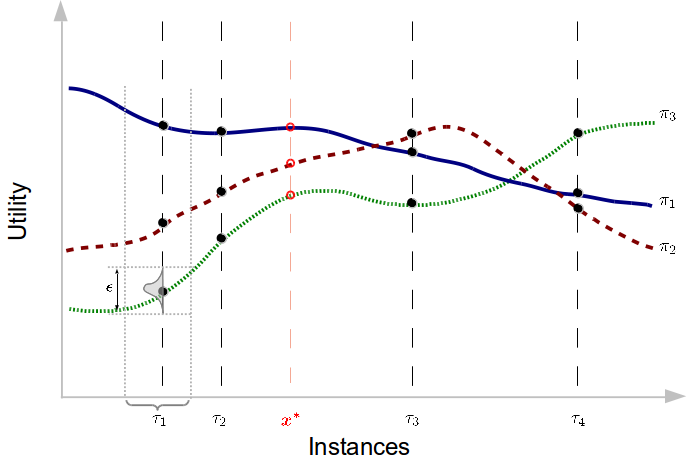}
\caption{A simplified depiction of the Bayesian Policy Reuse problem. The agent has access to a library of policies ($\pi_1$, $\pi_2$ and $\pi_3$), and has previously experienced a set of task instances ($\tau_1,\tau_2,\tau_3, \tau_4$), as well as samples of the utilities of the library policies on these instances (the black dots indicate the means of these estimates, while the agent maintains distributions of the utility as illustrated by $\mathrm{P}(U|\tau_1,\pi_3)$ in grey). The agent is presented with a new unknown task instance ($x^*$), and it is asked to select the best policy from the library (optimising between the red hollow points) without having to try every individual option (in less than 3 trials in this example). The agent has no knowledge about the complete curves, where the task instances occur in the problem space, or where the new task is located in comparison to previous tasks. This is inferred from utility similarity. For clarity, only performance is shown while the observation models are 
not depicted.\label{fig:bpr}}
\end{figure}

A caricature of the BPR problem for a  one-dimensional task space  is shown in Figure~\ref{fig:bpr}, where, given a new task $x^* \in \mathcal{X}$, the agent is required to select the best policy $\pi^* \in \Pi$ in as few trials as possible, whilst minimising the accumulated regret in the interim. As shown in this example, the agent has prior knowledge in the form of performance models for each policy in $\Pi$ on a set of tasks from $\mathcal{X}$. The agent additionally has observation models of the signals generated by each task-policy pair, but these are not depicted in Figure~\ref{fig:bpr}).

\subsection{General Algorithm}
Applying a specific policy on a new task instance generates sample observation signals, which are used along with the observation model to update a `similarity' measure over the previously-solved tasks (which is a distribution we call the \emph{belief}). This belief informs the selection of a policy at the next trial in an attempt to optimise expected performance. This is the core step in the operation of Bayesian Policy Reuse.

We present the general form of Bayesian Policy Reuse (BPR) in Algorithm~\ref{alg:bpr}. The policy selection step (line \ref{algline:reuse}) is described in detail in Section \ref{sec:selection}, the models of observation signals (line \ref{algline:obs}) are described in Section \ref{sec:models}, and the belief update (line \ref{algline:bayes}) is discussed further in Section \ref{sec:belief}.

\begin{algorithm}[!h]
\begin{algorithmic}[1]
\REQUIRE Problem space $\mathcal{X}$, Policy library $\Pi$, observation space $\Sigma$, prior over the problem space $\mathrm{P}(\mathcal{X})$, observation model $\mathrm{P}(\Sigma|\mathcal{X}, \Pi)$, performance model $\mathrm{P}(U|\mathcal{X}, \Pi)$, number of episodes $K$.
\STATE Initialise beliefs: $\beta^0(\mathcal{X}) \longleftarrow \mathrm{P}(\mathcal{X})$.
\FOR{episodes $t=1\ldots K$}
\STATE \label{algline:reuse} Select a policy $\pi^t \in \Pi$  using the current belief $\beta^{t-1}$ and the performance model $\mathrm{P}(U|\mathcal{X}, \pi^t)$.
\STATE Apply $\pi^t$ on the task instance.
\STATE \label{algline:obs} Obtain an observation signal $\sigma^t$ from the environment.
\STATE \label{algline:bayes} Update the belief $\beta^{t}(\mathcal{X}) \propto {\mathrm{P}(\sigma^t|\mathcal{X}, \pi^t)\beta^{t-1}(\mathcal{X})}$.
\ENDFOR
\end{algorithmic}
\caption{Bayesian Policy Reuse (BPR)}\label{alg:bpr}
\end{algorithm}

\subsection{Regret}
In order to evaluate the performance of our approach, we define \emph{regret} as the criterion for policy selection to be optimised by Bayesian Policy Reuse.
\begin{defm}[Library Regret]
For a library of policies $\Pi$ and for a policy selection algorithm $\xi: \mathcal{X}'\rightarrow \Pi$ that selects a policy for the new task instance $x^*\in\mathcal{X}'$, the \emph{library regret} of $\xi$ is defined as
\begin{equation*}
 \mathcal{R}_{\Pi}(\xi,x^*)=U_{x^*}^{\pi^*}-U_{x^*}^{\xi(x^*)},
\end{equation*}
where $U_{x}^{\pi}$ is the utility of policy $\pi$ when applied to task $x$, and $\pi^*=\arg \max_{\pi\in\Pi} U_{x^*}^\pi$, is the best policy in hindsight \emph{in the library} for the task instance $x^*$.
 
\end{defm}

\begin{defm}[Average Library Regret]
For a library of policies $\Pi$ and for a policy selection algorithm $\xi: \mathcal{X}'\rightarrow \Pi$, the \emph{average library regret} of $\xi$ over $K$ trials is defined as the average of the library regrets for the individual trials,

\begin{eqnarray*}
 \mathcal{R}_\Pi^K(\xi)&=&\frac{1}{K}\sum_{t=1}^K \mathcal{R}_\Pi(\xi,x^t),
\end{eqnarray*}
for a sequence of task instances $x^1, x^2, \ldots, x^K\in\mathcal{X}'$.
\end{defm}

The metric we minimise in  BPR is the average library regret $\mathcal{R}_\Pi^K(.)$ for $K$ trials. That is, the goal of BPR is to not only find the right solution at the end of the $K$ trials, possibly through expensive exploration, but also to optimise performance even when exploring in the small number of trials of the task. We will refer to this metric simply as `regret' throughout the rest of the paper.




\subsection{Types}
\label{sec:types}
When the problem space of BPR is a large task space $\mathcal{M}$, 
modelling the true belief distribution over the complete space would typically require a large number of samples (point estimates), which would hence be expensive to maintain and use computationally. 
In many applications, there is a natural notion of clustering in $\mathcal{M}$ whereby many tasks, modelled as MDPs, are similar with only minor variations in transition dynamics or reward structures. 
In the context of MDPs, previous work has regarded classes of MDPs as probability distributions over task parameters~\citep{mtrl}.   A more recent work explored explicitly discovering the clustering in a space of tasks~\citep{jmlr}.
Similar intuitions have been developed in the multi-armed bandits literature, by examining ways of clustering bandit machines in order to allow for faster convergence and better credit assignment, e.g. \cite{depndentbandits,clustered_bandits,latent_bandits}.
In this work we do not explicitly investigate methods of task clustering, but the algorithms presented herein are most efficient when such a cluster-based structure exists in the task space.

We encode the concept of task clustering by introducing a notion of task \emph{types} as $\epsilon$-balls in the space of tasks, where the tasks are clustered with respect to the performance of a collection of policies executed on them. 

\begin{defm}[Type]\label{defn:type}
A type $\tau$ is a subset of tasks such that for any two tasks $\mu_i,\mu_j$ from a single type $\tau$, and for all policies $\pi$ in a set of policies $\Pi$, the difference in utility is upper-bounded by some $\epsilon\in\mathds{R}$:
\begin{eqnarray*}\label{eqn:returnloss}
\mu_i,\mu_j\in\tau  \Leftrightarrow |U^\pi_i - U^\pi_j|\le\epsilon,\hspace{2mm} \forall \pi\in\Pi,
\end{eqnarray*}
where $U^\pi_i \in \mathds{R}$ is the utility from executing policy $\pi$ on task $\mu_i$. Then, $\mu_i$ and $\mu_j$ are \emph{$\epsilon$-equivalent} under the policies $\Pi$.
\end{defm}

This definition of types is similar to the concept of similarity-based contextual bandits~\citep{similarity}, where a distance function can be defined in the joint space of contexts and arms given by an upper bound of reward differences. In our setting, we cluster the instances (contexts) that are less than $\epsilon$-different under all the policies in the library (arms). We do not however assume any prior knowledge of the metrics in the task or policy spaces.

Figure~\ref{fig:bpr} depicted four example types, where each accounts for an $\epsilon$-ball in performance space (only explicitly shown for $\tau_1$). Note that the definition does not assume that the types need to be disjoint, i.e. there may exist tasks that belong to multiple types. We denote the space of types with $\mathcal{T}$. 

\begin{figure}[ht!]
\begin{center}
\includegraphics[width=0.9\textwidth]{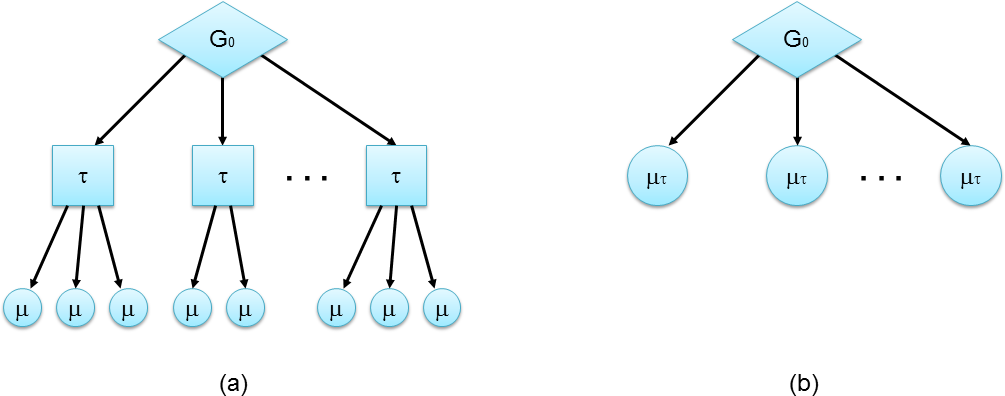}
\caption{Problem space abstraction model under disjoint types. (a) Tasks $\mu$ are related by types $\tau$, with a generating distribution $G_0$ over them. (b) A simplification of the hierarchical structure under $\epsilon$-equivalence. The tasks of each type are represented by a single task $\mu_\tau$.}
\label{fig:modelstructure}
\end{center}
\end{figure}

In the case of disjoint types, the type space $\mathcal{T}$ can be used as the problem space of BPR, inducing a hierarchical structure in the space  $\mathcal{M}$. The environment can then be represented with the generative model shown in Figure~\ref{fig:modelstructure}(a) where a type $\tau$ is drawn from a hyperprior $\tau \sim G_0$, and then a task is drawn from that type $\mu \sim \Delta^\tau(\mu)$, where $\Delta^\tau(.)$ is some probability distribution over the tasks of type $\tau$.

By definition, the set of MDPs generated by a single type are $\epsilon$-equivalent under $\Pi$, hence BPR regret  cannot be more than $\epsilon$ if we represent  all the MDPs in $\tau$ with any one of them. Let that chosen  MDP  be a \emph{landmark MDP} of type $\tau$, and denote this by $\mu_\tau$. This reduces the hierarchical structure  into the simpler model shown in Figure~\ref{fig:modelstructure}(b), where the prior acts immediately 
on a set of landmark MDPs $\mu_\tau$, $\tau\in\mathcal{T}$. The benefit of this for BPR is that each alternative is representative for a region in the original task space, as defined by a maximum loss of $\epsilon$. Maintaining only this reduced set of landmarks removes near-duplicate tasks from consideration, thereby reducing the cost of maintaining the belief. 

For the remainder of this paper, we use the type space $\mathcal{T}$ as the problem space, although we note that the methods proposed herein do not prevent the alternative use of the full task space $\mathcal{M}$.

\subsection{Performance Model}
\label{sec:perfmodel}
One of the key components of BPR is the performance model of policies on previously-solved task instances, which describes the distribution of returns from each policy on the previously-solved tasks. A performance model represents the variability in return under the various tasks in a type.

\begin{defm}[Performance Model]
For a policy $\pi$ and a type $\tau$, the \emph{performance model} $\mathrm{P}(U|\tau,\pi)$ is a probability distribution over the utility of $\pi$ when applied to all tasks $\mu\in\tau$.
 \end{defm}

Figure~\ref{fig:bpr} depicts the performance profile for $\pi_3$ on type $\tau_1$ in the form of a Gaussian distribution. Recall that for a single type, and each policy, the domain of the performance model would be at most of size $\epsilon$. The agent maintains performance models for all the policies it has in the library $\Pi$ and for all the types it has experienced.



\subsection{Observation Model}
\label{sec:models}
In order to facilitate the task identification process, the agent learns a model of how types, policies and signals relate during the offline phase. 

\begin{defm}[Observation Model]
 For a policy $\pi$ and type $\tau$ and for a choice of signal space $\Sigma$, the \emph{observation model} $\mathcal{F}_{\pi}^{\tau}(\sigma) = \mathrm{P}(\sigma|\tau, \pi)$ is a probability distribution over the signals $\sigma\in\Sigma$ that may result by applying the policy $\pi$ to the type $\tau$.
\end{defm}
We consider the following offline procedure to learn the signal models for a policy library $\Pi$:
\begin{enumerate}
 \item The type label $\tau$ is announced.
  \item A set of tasks are generated from the type $\tau$.
 \item The agent runs all the policies from the library $\Pi$ on all the instances of $\tau$, and observes the resultant sampled signals  $\tilde{\sigma}\in\Sigma$.
 \item Empirical distributions $\mathcal{F}_{\pi}^{\tau}=\Delta(\tilde{\sigma})$ are fitted to the data, for each type $\tau$ and policy $\pi$.
\end{enumerate}


The benefit of these models is that they provide a connection between the observable online information and the latent type label, 
the identification of which leads to better reuse from the  policy library.

\subsection{Candidate Signals for Observation Models}
\label{sec:signals}

The BPR algorithm requires that some signal information is generated from policy execution on a task, although the form of this signal remains unspecified. Here we describe the three most typical examples of information that can be used as signals in BPR, but note that this list is not exhaustive.

\subsubsection{State-Action-State Tuples}
The most detailed information signal which could be accrued by the agent is the history of all $(s,a,s')$ tuples encountered during the execution of a policy. Thus, the observation model in this case is an empirical estimate of the expected transition function of the MDPs under the type $\tau$.

The expressiveness of this signal does have a drawback, in that it is expensive to learn and maintain these models for every possible type. Additionally, this may not generalise well, in cases with sparse sampling. 
On the other hand, this form of signal is useful in cases where some environmental factors may affect the behaviour of the agent in a way that does not directly relate to attaining an episodic goal. As an example, consider an aerial agent which may employ different navigation strategies under different wind conditions.

\subsubsection{Instantaneous Rewards}
Another form of information is the instantaneous reward $r \in \mathds{R}$ received during the execution of a policy for some state-action pair. Then, the observation model is an empirical estimate of the expected reward function for the MDPs in the type.

Although this is a more abstract signal than the state-action-state tuples, it may still provide a relatively fine-grained knowledge on the behaviour of the task when intermediate rewards are informative. It is likely to be useful in scenarios where the task has a number of subcomponents which individually contribute to overall performance, for example in assembly tasks.

\subsubsection{Episodic Returns}
An example of a sparser kind of signal is the total utility $U^\pi_\tau \in \mathds{R}$ accrued over the full episode of using a policy in a task. The observation model of such a scalar signal is much more compact, and thereby easier to learn and reason with, than the previous two proposals. We also note that for our envisioned applications, the execution of a policy cannot be terminated prematurely, meaning that an episodic return signal is always available to the agent before selecting a new policy.

This signal is useful for problems of delayed reward, where intermediate states cannot be valued easily, but the extent to which the task was successfully completed defines the return. In our framework, using episodic returns as signals has the additional advantage that this information is already captured in the performance model, which relieves the agent from maintaining two separate models, as in this case $\mathrm{P}(U|\tau, \pi) = \mathcal{F}_{\pi}^{\tau}(U)$ for all $\pi$ and $\tau$.


\subsection{Belief over Types}
\label{sec:belief}
\begin{defm}[Type Belief]
For a set of previously-solved types $\mathcal{T}$ and a new instance $x^*$, the \emph{Type Belief} $\beta(.)$ is a probability distribution over $\mathcal{T}$ that measures the extent to which $x^*$ matches the types of $\mathcal{T}$ in their observation signals.

\end{defm}

The type belief, or \emph{belief} for short, is a surrogate measure of similarity in type space. It approximates where a new instance may be located in relation to the known types  which act as a basis of the unknown type space.  The belief is initialised with the prior probability over the types, labelled $G_0$ in Figure~\ref{fig:modelstructure}.

In episode $t$, the environment provides an observation signal $\sigma^t$  for executing a policy $\pi^t$ on the new task instance. This signal is used to update $\beta$ (line~\ref{algline:bayes} in Algorithm~\ref{alg:bpr}). The posterior over the task space 
is computed using Bayes' rule:
\begin{eqnarray}\label{eqn:bayes}
 \beta^{t}(\tau) & = & \frac{\mathrm{P}(\sigma^t|\tau, \pi^t) \beta^{t-1}(\tau)}{\sum_{\tau' \in \mathcal{T}} \mathrm{P}(\sigma^t|\tau', \pi^t) \beta^{t-1}(\tau')}\\
& = & \eta\,{\mathcal{F}_{\pi^t}^\tau(\sigma^t) \beta^{t-1}(\tau)},\hspace{4mm}  \forall \tau\in\mathcal{T},
\end{eqnarray}
where $\beta^{t-1}$ is the belief after episode $t-1$ and $\eta$ is a normalisation constant. We use $\beta$ to refer to $\beta^t$ whenever this is not ambiguous. Note how the belief is updated using the observation model.








\section{Policy Selection for BPR}
\label{sec:selection}
The selection of a policy for each episode (line~\ref{algline:reuse} in Algorithm~\ref{alg:bpr}) is a critical step in BPR. Given the current type belief, the agent is required to choose a policy for the next episode to fulfil two concurrent purposes: acquire useful information about the new (current) task instance, and at the same time avoid accumulating additional regret. 

At the core of this policy selection problem is the trade-off between exploration and exploitation. When a policy is executed at some time $t$, the agent receives both some utility as well as information about the true type of the new task (the signal). The agent is required to gain as much information about the task as possible, so as to choose policies optimally in the future, but at the same time minimise performance losses arising from sub-optimal policy choices\footnote{Note that we denote by \emph{optimal policy} the best policy in the library for a specific instance, as we are considering policy reuse problems in which learning the actual optimal policy is not feasible.}. 


Our problem can be mapped to a  finite-horizon total-reward multi-armed bandits setting in which the arms represent the policies, the finite horizon is defined by the limited number of episodes, and the metric to optimise is the total reward. For this kind of setting~\cite{lai1985asymptotically} show that index-based methods  achieve  optimal performance asymptotically. In our case, however, we are interested in the cumulative performance over a small number of episodes.

Clearly, a purely greedy policy selection mechanism would fail to choose exploratory options to elicit what is needed for the belief to converge  to the closest type
, and may result in the agent becoming trapped in a local maximum of the utility function. On the other hand, a purely exploratory policy selection mechanism could be designed to ensure that all the possible information is elicited in expectation, but this would not make an effort to improve performance instantly and thereby incur additional regret.  We thus require a mechanism to explore as well as exploit; find a better policy to maximise asymptotic utility, and exploit the current estimates of which are good policies to maximise myopic utility.

Multiple proposals have been widely considered in the multi-armed bandits (MAB) literature for these heuristics, ranging from early examples like the Gittins index for infinite horizon problems~\citep{gittins} to more recent methods such as the knowledge gradient~\citep{knowgrad}. Here we describe several approximate policy selection mechanisms that we use for dealing with the policy reuse problem.

\begin{itemize}

\item A first approach is through \emph{$\epsilon$-greedy exploration}, where with probability $1-\epsilon$ we select the policy which maximises the expected utility under the belief $\beta$, 
\begin{eqnarray*}
\hat{\pi}&=&\arg \max_{\pi\in\Pi}\sum_{\tau\in\mathcal{T}}\beta(\tau)\int_{U\in\mathds{R}}U\hspace{1mm}\mathrm{P}(U|\tau,\pi)\mathrm{d}U \vspace{6mm}\\
&=&\arg \max_{\pi\in\Pi}\sum_{\tau\in\mathcal{T}}\beta(\tau)\;\mathrm{E}[U|\tau,\pi],
\end{eqnarray*}
and with probability $\epsilon$ we select a policy from the policy library uniformly at random. This additional random exploration component perturbs the belief from local minima.

\item A second approach is through \emph{sampling the belief} $\beta$. This involves sampling a type according to its probability in the belief $\hat{\tau}\sim\beta$, and playing the best response to that type from the policy library,
\begin{equation*}\hat{\pi}=\arg \max_{\pi\in\Pi}\mathrm{E}[U|\hat{\tau},\pi].\end{equation*}

In this case, the sampled type acts as an approximation of the true unknown type, and exploration is achieved through the sampling process.

\item The third approach is through employing what we call \emph{exploration heuristics}, which are functions that estimate a value for each policy which measures the extent to which it balances exploitation with a limited degree of look-ahead for exploration.


This is the prevalent approach  in Bayesian optimisation, where, instead of directly maximising the objective function itself (here, utility), a surrogate function that takes into account both the expected utility and a notion of the utility variance (uncertainty)  is maximised (see, \emph{e.g}.,~\cite{bayesopt}). 
independent of other policies.


\end{itemize}

\subsection{Bayesian Policy Reuse with Exploration Heuristics}
\label{sec:bprkg}

By incorporating the notion of 
an exploration heuristic that computes an index $\nu_\pi$ for a policy $\pi$ into Algorithm~\ref{alg:bpr}, we obtain the proto-algorithm Bayesian Policy Reuse with Exploration Heuristics (BPR-EH) 
 described in Algorithm~\ref{alg:bprkg}.

\begin{algorithm}[!h]
\begin{algorithmic}[1]
\REQUIRE Type space $\mathcal{T}$, Policy library $\Pi$, observation space $\Sigma$, prior over the type space $G_0$, observation model $\mathrm{P}(\Sigma|\mathcal{T}, \Pi)$, performance model $\mathrm{P}(U|\mathcal{T}, \Pi)$, number of episodes $K$, exploration heuristic $\mathcal{V}$.
\STATE Initialise beliefs: $\beta^0 \longleftarrow G_0.$\label{ln:init}
\FOR{episodes $t=1\ldots K$}
\STATE \label{ln:compute_heuristic}Compute $\nu_\pi=\mathcal{V}(\pi, \beta^{t-1})$ for all 
 $\pi \in \Pi$\label{ln:eh}. 
\STATE $\pi^t \longleftarrow \arg \max_{\pi \in \Pi} \nu_\pi$.
\STATE Apply $\pi^t$ to the task instance.
\STATE Obtain the observation signal $\sigma^t$ from the environment.
\STATE Update the belief $\beta^{t}$ using $\sigma^t$ by Equation~(\ref{eqn:bayes}).
\ENDFOR
\end{algorithmic}
\caption{Bayesian Policy Reuse with Exploration Heuristics (BPR-EH)\label{alg:bprkg}}
\end{algorithm}

Note that we are now using $G_0$, the hyper-prior, as the prior in line~\ref{ln:init} because we are using  $\mathcal{T}$ as the problem space. We now define the exploration heuristics $\mathcal{V}$ that are used in line~\ref{ln:compute_heuristic}, and to this end we define four variants of the BPR-EH algorithm, as
\begin{itemize}
 \item BPR-PI using \emph{probability of improvement} (Section~\ref{sec:piei}),
 \item BPR-EI using \emph{expected improvement} (Section~\ref{sec:piei}),
 \item BPR-BE using \emph{belief entropy} (Section~\ref{sec:be}), and
 \item BPR-KG using \emph{knowledge gradient} (Section~\ref{sec:kg}).
\end{itemize}



\subsubsection{Probability of Improvement and Expected Improvement}
\label{sec:piei}

The first heuristic for policy selection utilises the probability with which a specific policy can achieve a hypothesised increase in performance. Assume that $U^{+}\in\mathds{R}$ is some utility which is larger than the best estimate under the current belief, $U^{+}>\bar{U}=\max_{\pi\in\Pi}\sum_{\tau\in\mathcal{T}}\beta(\tau)\mathrm{E}[U|\tau,\pi]$. The \emph{probability of improvement} (PI) principle chooses the policy that maximises the term,
\begin{equation*}
 \hat{\pi}=\arg \max_{\pi\in\Pi}\sum_{\tau\in\mathcal{T}}\beta(\tau) \mathrm{P}(U^+|\tau,\pi),
\end{equation*}
thereby selecting the policy most likely to achieve the utility $U^+$.

The choice of $U^+$ is not straightforward, and this choice is the primary factor affecting the performance of this exploration principle. One approach to addressing this choice, is through the related idea of \emph{expected improvement} (EI). This exploration heuristic integrates over all the possible values of improvement $\bar{U}<U^+<U^{max}$, and the policy is chosen with respect to the best potential. That is,
\begin{eqnarray*}
 \hat{\pi}&=&\arg \max_{\pi\in\Pi}\int_{\bar{U}}^{U^{max}}\sum_{\tau\in\mathcal{T}}\beta(\tau) \mathrm{P}(U^+|\tau,\pi)\mathrm{d}U^+\\
&=&\arg \max_{\pi\in\Pi}\sum_{\tau\in\mathcal{T}}\beta(\tau)\int_{\bar{U}}^{U^{max}} \mathrm{P}(U^+|\tau,\pi)\mathrm{d}U^+\\
&=&\arg \max_{\pi\in\Pi}\sum_{\tau\in\mathcal{T}}\beta(\tau)(1-\mathrm{F}(\bar{U}|\tau,\pi))\\
&=&\arg \min_{\pi\in\Pi}\sum_{\tau\in\mathcal{T}}\beta(\tau)\mathrm{F}(\bar{U}|\tau,\pi),
\end{eqnarray*}

where $\mathrm{F}(U|\tau,\pi)=\int_{-\infty}^U \mathrm{P}(u|\tau,\pi)\mathrm{d}u$ is the cumulative distribution function of $U$ for $\pi$ and $\tau$. This heuristic therefore selects the policy most likely to result in any improvement to the expected utility.

\subsubsection{Belief Entropy}
\label{sec:be}

Both PI and EI principles select a policy which has the potential to achieve higher utility. An alternate approach is to select the policy which will have the greatest effect in reducing the uncertainty over the type space.

The \emph{belief entropy} (BE) exploration heuristic seeks to estimate the effect of each policy in reducing uncertainty over type space, represented by the entropy of the belief. 
 For each policy $\pi \in \Pi$, estimate the expected entropy of the belief after executing $\pi$ as
\begin{eqnarray*}
  \mathrm{H}(\beta|\pi) & = & -\beta^\pi \log \beta^\pi,
\end{eqnarray*}
where $\beta^\pi$ is the updated belief after seeing  the  signal expected from running $\pi$, given as
\begin{eqnarray}
 \beta^\pi(\tau) & = & \mathrm{E}_{\sigma\in\Sigma}\left[ {\eta\,\mathcal{F}^\tau_\pi(\sigma) \beta(\tau)}
\right]\\
 & = & \int_{\sigma \in \Sigma} \mathcal{F}_\pi^{\beta}(\sigma) \left[\eta\,{\mathcal{F}^\tau_\pi(\sigma) \beta(\tau)}\right] \mathrm{d}\sigma,\label{eqn:updatbeta}
\end{eqnarray}
where $\mathcal{F}_\pi^{\beta}(\sigma)$ is the probability of observing $\sigma$ under the current belief $\beta$ when using $\pi$, and $\eta$ is the normalisation constant as before.

Then, selecting the policy
\begin{equation*}
 \hat{\pi}=\arg \min_{\pi\in\Pi} \mathrm{H}(\beta|\pi)
\end{equation*}
 reduces the most uncertainty in the belief in expectation. This is however a purely exploratory policy. To incorporate exploitation of the current state of knowledge, we rather select 
\begin{equation*}
 \hat{\pi}=\arg \max_{\pi\in\Pi} \left(\tilde{U}(\pi) - \kappa \mathrm{H}(\beta|\pi)\right),
\end{equation*}
where $\kappa\in\mathds{R}$ is a positive constant controlling the exploration-exploitation trade-off, and $\tilde{U}(\pi)$ is the expected utility of $\pi$ under the current belief,  
\begin{eqnarray}\label{eqn:expectedutility}
 \tilde{U}(\pi) 
 & = & \sum_{\tau \in \mathcal{T}} \beta(\tau) \mathrm{E}[U|\pi,\tau].
\end{eqnarray}

\subsubsection{Knowledge Gradient}
\label{sec:kg}

The final exploration heuristic we describe is the \emph{knowledge gradient}~\citep{knowgrad}, which aims to balance exploration and exploitation through optimising myopic return whilst maintaining asymptotic optimality. The principle behind this approach is to estimate a one step look-ahead, and select the policy which maximises utility over both the current time step  and the next in terms of the information gained. 

To select a policy using the knowledge gradient, we choose the policy which maximises the online knowledge gradient at time $t$
\begin{eqnarray*}
 \hat{\pi} & = & \arg \max_{\pi \in \Pi} \left(\tilde{U}(\pi) + (K-t)\nu_\pi^{t}\right),
\end{eqnarray*}
trading-off between the expected utility $\tilde{U}(\pi)$, given in Equation~\ref{eqn:expectedutility}, and $\nu_\pi^{t}$, the offline knowledge gradient of $\pi$ for a horizon of $K$ trials, weighted by the remaining number of trials. The offline knowledge gradient essentially measures the performance of a one-step look-ahead in the process, given as 
\begin{eqnarray}\label{eqn:kg}
\nu_\pi^{t} & = & \mathrm{E}_\beta\left[\max_{\pi'} \tilde{U}^{\pi}(\pi')  - \max_{\pi''} \tilde{U}(\pi'') \right],
\end{eqnarray}
where  $\tilde{U}(\pi)$ is the expected utility of $\pi$ under the current belief (Equation~\ref{eqn:expectedutility}), 
\begin{eqnarray}
\tilde{U}^{\pi}(\pi')=\sum_{\tau\in\mathcal{T}}\beta^\pi(\tau)\;\mathrm{E}[U|\tau,\pi'],
\end{eqnarray}
and $\beta^\pi$ is the expected updated belief after playing policy $\pi$ and receiving a suitable signal, as defined in Equation~\ref{eqn:updatbeta}. 
That is, the offline knowledge gradient is the difference in expectation, with respect to $\beta$, of the best performance of any policy at $t+1$ if $\pi$ was played at $t$, compared to that of the best policy at $t$ (which may be different from $\pi$).


\section{Experiments}

\subsection{Golf Club Selection}

As an initial, illustrative simulated experiment we consider the problem of a robot golfer taking a shot with one of four golf clubs on an unknown 
 golf course, where it is \emph{not possible} to reliably estimate the distance to the hole, as for this example we are considering a robot with weak sensors that are not in themselves sufficient to reliably measure distance. The robot is only allowed to take $K=3$ shots, which is less than the number of available clubs, from a fixed position from the hole. The task is evaluated by the stopping distance of the ball to the hole. The robot can choose any of the available clubs, and we assume that the robot uses a fixed, canonical stroke with each club.

In this setting, we consider the type space $\mathcal{T}$ to be a set of different golfing experiences the robot had before, each defined for simplicity by how far the target was  (other factors, e.g. weather conditions, could be factored into this as well). The performance of a club for some hole is defined as the negative of the 
absolute distance of the end location of the ball from the hole, such that this quantity must be maximised. 

Then, 
the choice of a club corresponds to a choice of a policy. For each, the robot  has a performance profile (distribution over final distance of the ball from the hole) for the different courses that the robot experienced. 
 We assume a small selection of four clubs, with properties shown in Table~\ref{tab:clubs} for the robot canonical stroke. The distances shown in this table are the ground truth values,  and are not explicitly known to the robot.

\begin{table}[h]
\begin{center}
\begin{tabular}{l|cc} 
Club & Average Yardage & Standard Deviation of Yardage\\ \hline
$\pi_1=$ 3-wood & 215 & 8.0 \\
$\pi_2=$ 3-iron & 180 & 7.2 \\
$\pi_3=$ 6-iron & 150 & 6.0 \\
$\pi_4=$ 9-iron & 115 & 4.4
\end{tabular}
\caption{Statistics of the ranges (yardage) of the four clubs used in the golf club selection experiment. We choose to model the performance of each club by a Gaussian distribution. We assume the robot is competent with each club, and so the standard deviation is small, but related to the distance hit.\label{tab:clubs}}
\end{center}
\end{table}

\begin{figure}[!h]
\begin{center}
\includegraphics[width=0.8\textwidth]{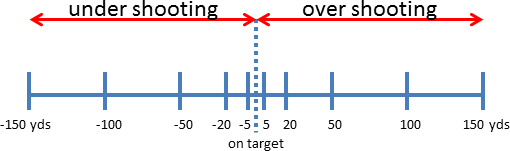}
\caption{The distance ranges used to provide observation signals. The robot is only able to identify which of these bins corresponds to the end location of the ball after a swing.}
\label{fig:golfranges}
\end{center}
\end{figure}

Owing to the difficulty of the outdoor perception problem over large distances, the robot cannot measure exact distances in the field, but for a feedback signal, it can crudely estimate a qualitative description of the result of a shot as falling into one of several broad categories (corresponding to concepts such as \emph{quite near} and \emph{very far}), which define the observation space, as shown in Figure \ref{fig:golfranges}.

\begin{figure}[!h]
\begin{center}
\includegraphics[width=\textwidth]{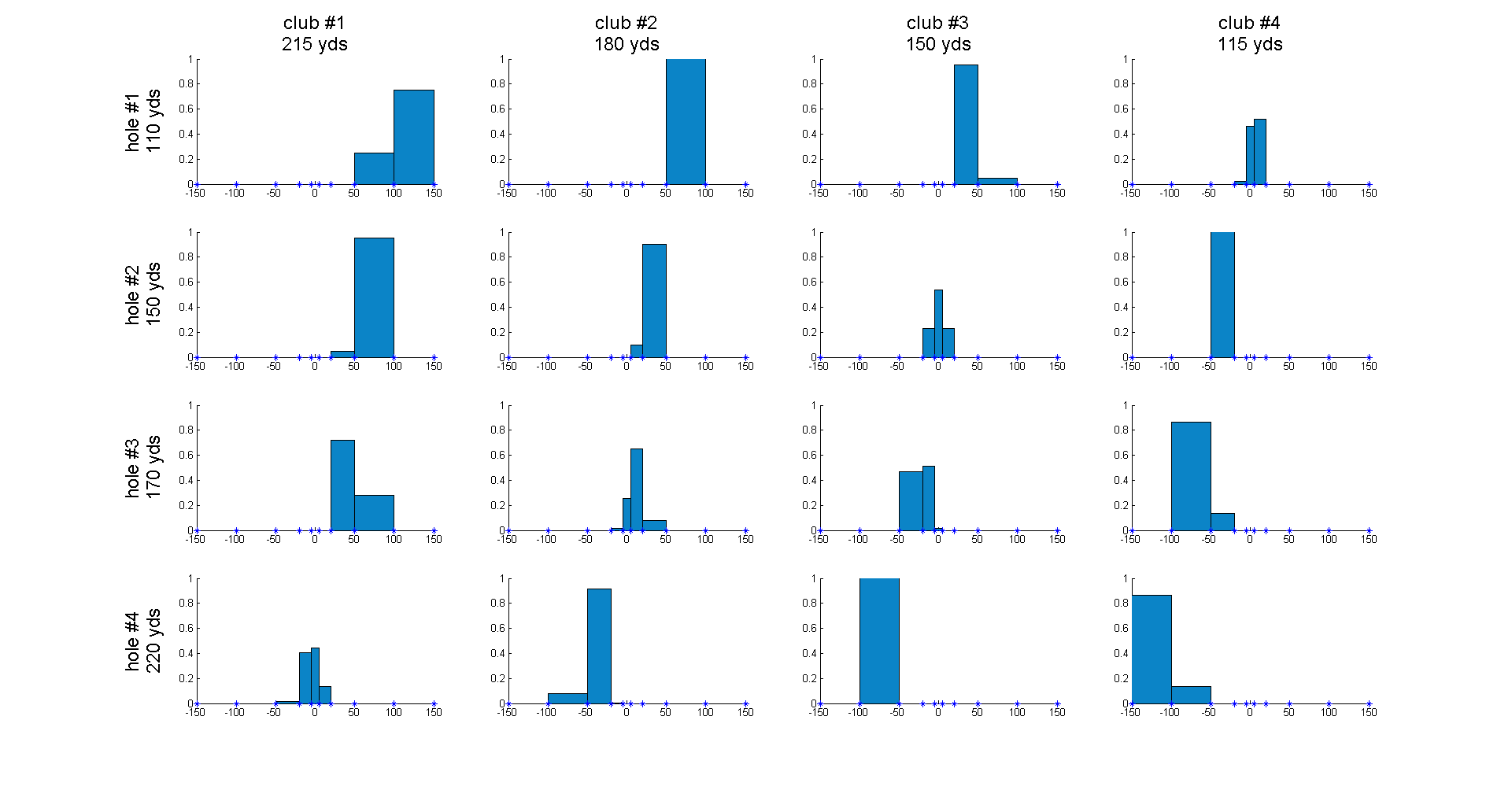}
\caption{Performance-correlated signal models for the four golf clubs in Table~\ref{tab:clubs} on four training holes with distances 110yds, 150yds, 170yds and 220yds. The models capture the probabilities of the ball landing in the corresponding distance categories. The width of each category bin has been scaled to reflect the distance range it signifies. The x-axis is the distance to the hole, such that negative values indicate under-shooting, and positive distances over-shooting the hole.}
\label{fig:golfperf}
\end{center}
\end{figure}

Note that this is not the performance itself, but a weaker observation correlated with performance. The distributions over these qualitative categories (the observation models) are known to the robot for each club on each of the training types it has encountered. For this example, we assume the robot has extensive training on four particular holes, with distances $\tau_{110} =$ 110yds, $\tau_{150} =$ 150yds, $\tau_{170} =$ 170yds and $\tau_{220} =$ 220yds. The observation models are shown in Figure~\ref{fig:golfperf}.

When the robot faces a new hole, BPR allows the robot to overcome its inability to judge the distance to the hole by using the feedback from an arbitrary shot as a signal. The feedback signal updates an estimate of the most similar previous task (the belief), using the distributions in Figure~\ref{fig:golfperf}. This belief enables the robot to choose the club/clubs which would have been the best choice for the most similar previous task/tasks. 

For a worked example, consider a hole 179 yards away. If a coarse estimate of the distance is feasible, it can be incorporated as a prior over $\mathcal{T}$. Otherwise, an uniformed prior is used. Assume the robot is using greedy policy selection, and assume that it selects $\pi_1$ for the first shot due to a uniform prior, and that this resulted in an over-shot by 35 yards. The robot cannot gauge this error more accurately than that it falls into the category corresponding to {`over-shooting in the range of 20 to 50 yards'}. This signal will update the belief of the robot over the four types, and by Figure~\ref{fig:golfperf}, the closest type to produce such a behaviour would be $\tau_{170}=170$ yards. The new belief dictates that the best club to use for anything like $\tau_{170}$ is $\pi_2$. Using $\pi_2$, the hole is over-shot by 13 yards, corresponding to the category with the `range 5 to 20 yards'. Using the same calculation, the most similar previous type is again $\tau_{170}$, keeping the best club 
as $\pi_2$, and allowing belief to converge. Indeed, given the ground truth in Table~\ref{tab:clubs}, this is the best choice for the 179 yard task. Table~\ref{tab:golfworked} 
describes this process over the course of 8 consecutive shots taken by the robot.

\begin{table}[!h]
\begin{center}
\begin{tabular}{|l|cccccccc|}\hline
Shot & 1 & 2 & 3 & 4 & 5 & 6 & 7 & 8\\
\hline
Club & 1 & 2 & 2 & 2 & 2 & 2 & 2 & 2\\
Error & 35.3657 & 13.1603 & 4.2821 & 6.7768 & 2.0744 & 11.0469 & 8.1516 & 2.4527\\
Signal & 20--50 & 5--20 & -5--5 & 5--20 & -5--5 & 5--20 & 5--20 & -5--5\\
$\beta$ entropy & 1.3863 & 0.2237 & 0.0000 & 0.0000 & 0.0000 & 0.0000 & 0.0000 & 0.0000\\
\hline
\end{tabular}
\caption{The 179 yard example. For each of 8 consecutive shots: the choice of club, the true error in distance to the hole (in yards), the coarse category within which this error lies (the signal received by the agent), and the entropy of the belief. This shows convergence after the third shot, although the correct club was used from the second shot onwards. The oscillating error is a result of the variance in the club yardage. Although the task length was $K=3$ strokes, we show these results for longer to illustrate convergence. \label{tab:golfworked}}
\end{center}
\end{table}

\begin{figure}[!h]
\begin{center}
\includegraphics[width=\textwidth,height=0.27\textheight]{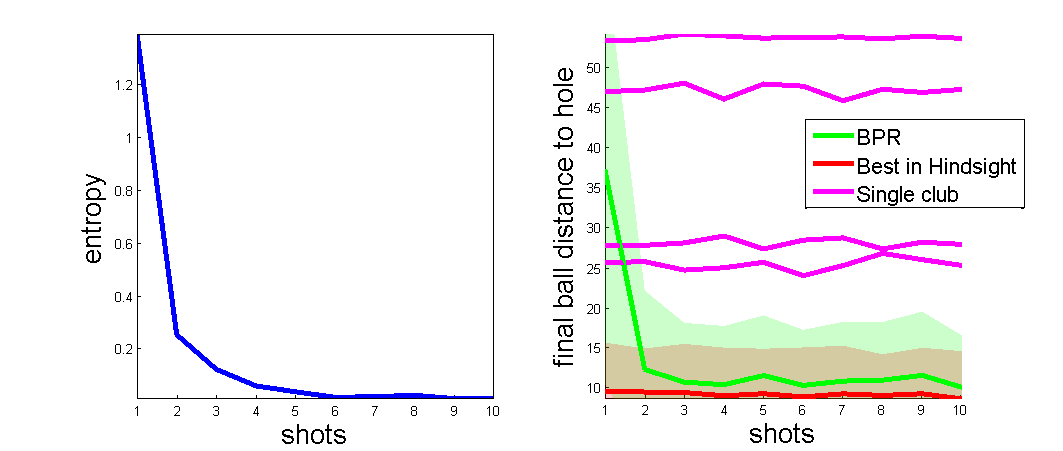}
\caption{Performance of BPR on the golf club example, with results averaged over 100 unknown holes, showing the decrease in entropy of the belief $\beta$ and average distance to the hole (lower is better). Performance of the pure four clubs (average performance of each single club over all 100 holes), as well as the best club for each hole in retrospect, is shown for regret comparison. Although the task length is $K=3$ strokes, we show these results for longer to illustrate convergence. Shaded regions denote one standard deviation. Error bars on the individual clubs have been omitted for clarity, but their average standard deviations are $26.85$m, $15.37$m, $20.89$m, and $27.66$m respectively.}
\label{fig:golfexample}
\end{center}
\end{figure}

Figure~\ref{fig:golfexample} shows the performance of BPR with greedy policy selection in the golf club selection task averaged over 100 unknown golf course holes, with ranges randomly selected between 120 and 220 yards. This shows that on average, by the second shot, the robot will have selected a club capable of bringing the ball within 10--15 yards of the hole.

\subsection{Online Personalisation}
\label{sec:exp:personalisation}

In this next experiment, we demonstrate the use of different observation signals to update beliefs, as described in Section~\ref{sec:signals}.

Consider an automated phone service of a bank, where the bank tries to improve the speed of answering telephonic queries using human operators by a personalised model for understanding the speech of the user and responding with a synthesised voice personalised according to that user's preferences. In a traditional speech recognition system, the user may have time to train the system to her own voice, but this is not possible in this scenario. As a result, the phone service may have a number of different pre-trained language models and responses, and over the course of many interactions with the same user, tries to estimate the best such model to use.

Let a user $i$ be defined by a preference model of language,  $\lambda_i \in \{ 1, \ldots, L \}$, where $L$ is the number of such models. The policy $\pi$ executed by the telephonic agent also corresponds to a choice of language model, i.e. $\pi \in \{ 1, \ldots, L \}$. The goal of the agent is to identify the user preference $\lambda_i$, whilst minimising frustration to the user.

\begin{figure}[!h]
\begin{center}
\includegraphics[width=0.7\textwidth]{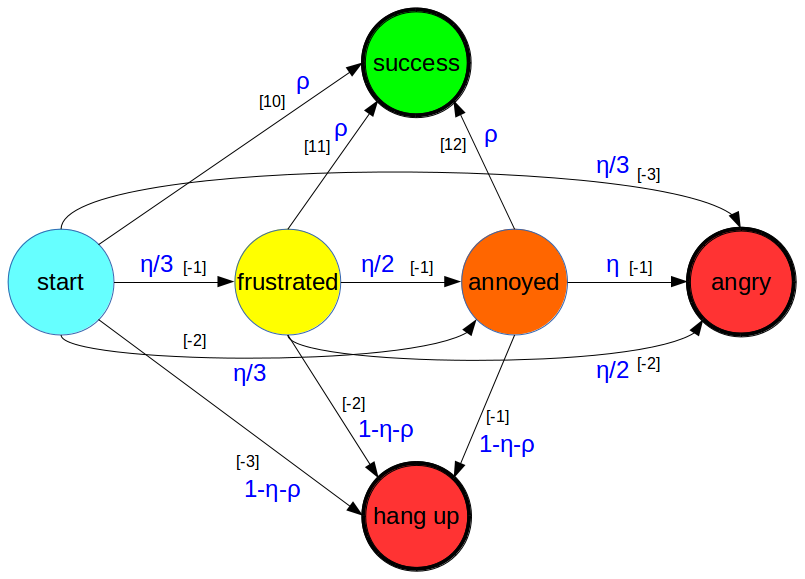}
\caption{Transition system describing the online telephonic personalisation example. Circles are states, thick bordered circles are terminal states, small black edge labels in square brackets are the transition rewards, and large blue edge labels are transition probabilities. See text for a description of the parameters $\rho$ and $\eta$.}
\label{fig:callmodel}
\end{center}
\end{figure}

Assume that each telephonic interaction proceeds using the transition system through the six states given in Figure~\ref{fig:callmodel}. In every state, there is only one action which can be taken by the system, being to use the chosen language model. At the beginning of the call, the user is in the \emph{start} state. We assume the system can identify the user by caller ID, and selects a language model. If, at any point, the system can deal with the user's request, the call ends successfully. If not, we assume the user becomes gradually more irritated with the system, passing through states \emph{frustrated} and \emph{annoyed}. If the user reaches state \emph{angry} and still has an unresolved request, she is transferred to a human operator. This counts as an unsuccessful interaction. Alternatively, at any point the user may hang up the call, which also terminates the interaction unsuccessfully.

The transition dynamics of this problem depend on a parameter $\rho = 1-\frac{|\pi - \lambda_i|}{L}$ which describes how well the selected language model $\pi$ can be understood by a user of type $\lambda_i$, such that $\rho =1$ if the chosen model matches the user's, and it is 0 if it is the worst possible choice. An additional parameter $\eta$ governs the trade-off between the user becoming gradually more frustrated  and  simply hanging up when the system does not respond as expected. In our experiments, we fix $\eta = 0.3$, except when $\pi = \lambda_i$ where we set $\eta = 0$.

To allow the use of different observation signals in this example the domain was designed in a way such that the transition dynamics and the rewards of this domain, shown in Figure~\ref{fig:callmodel}, allow only two total utility values for any instance: $U = 10$ for a successful completion of the task, and $U = -3$ otherwise. Similarly, any state that transitions to the unsuccessful outcome \emph{angry} receives the same reward for a transition to the unsuccessful outcome \emph{hang up}. Finally, all transition probabilities between the states \emph{start}, \emph{frustrated}, \emph{annoyed}, and \emph{angry} are independent of $\rho$, and thus, of the type. This set up mirrors the fact that in general the state sequence given by the signal $(s,a,s')$ is more informative than the reward sequence $(s,a,r)$, which is in turn more informative than the total utility signal $U$ alternative~\footnote{We note that in many 
applications $U$ might be the only of these signals available to the agent. For example, in the current scenario, it may not be easy or feasible to accurately gauge the frustration of the caller, making the states and the immediate rewards unobservable.}.

\begin{figure}[!h]
\begin{center}
\includegraphics[width=\textwidth]{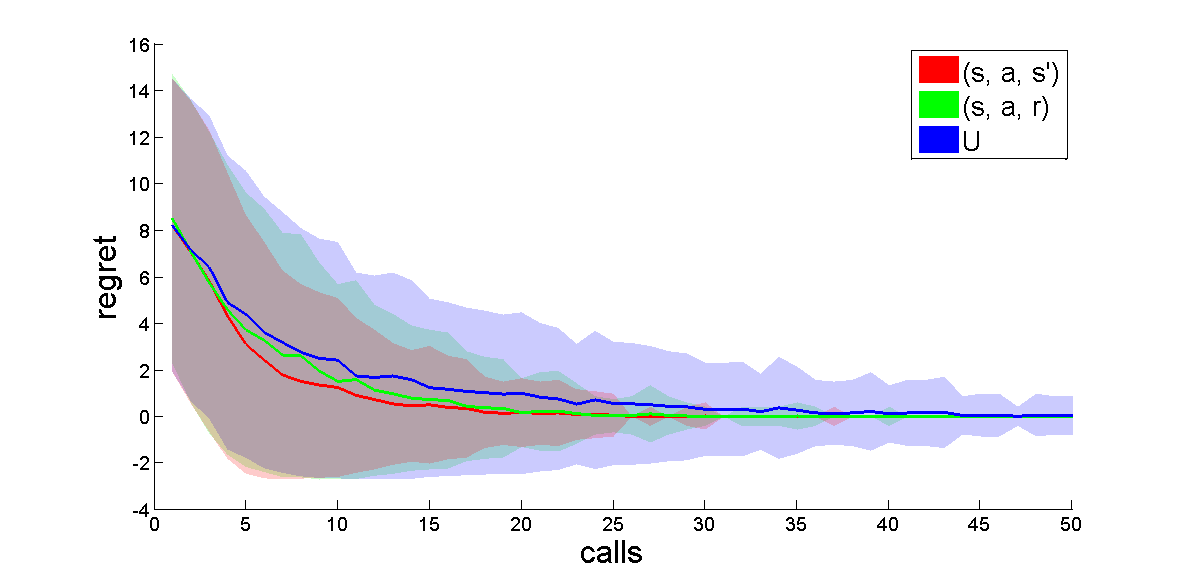}
\caption{Regret, showing comparative performance of BPR on the telephone banking domain, using $(s,a,s')$, $(s,a,r)$, and $U$ as signals.}
\label{fig:sassarr}
\end{center}
\end{figure}

The results shown in Figure~\ref{fig:sassarr} were generated from 1,000 call interactions which proceeded according to the model in Figure~\ref{fig:callmodel}. In this experiment, the correct language model for each user was randomly drawn from a set of 20 language models. Figure~\ref{fig:sassarr} shows comparative performance of BPR with \emph{sampling the belief} selection mechanism when the three kinds of signals are used. As expected, the lowest regret (and variance in regret) is achieved using the most-informative $(s,a,s')$ signal, followed by the $(s,a,r)$ signal, and finally the total performance signal $U$. We do note, however, that all three signals eventually converge to zero regret if given enough time.

\subsection{Surveillance Domain}
\label{sec:exp:surveillance}


The surveillance domain models the monitoring problem laid out in the introduction. Assume a base station is tasked with monitoring a wildlife reserve spread out over some large geographical region. The reserve suffers from poaching and so the base station is required to detect and respond to poachers on the ground. The base station has a fixed location, and so it monitors the region by deploying a low-flying, 
light-weight autonomous drone to complete particular surveillance tasks using different strategies. The episodic commands issued by the base station may be to deploy to a specific location, scan for unusual activity in the targeted area and then report back. After completing each episode, the drone communicates with the base some information of whether or not there was any suspicious activity in the designated region. The base station is required to use that information to better decide on the next strategy for the drone.

\begin{figure}[!h]
\begin{center}
\includegraphics[width=0.75\textwidth]{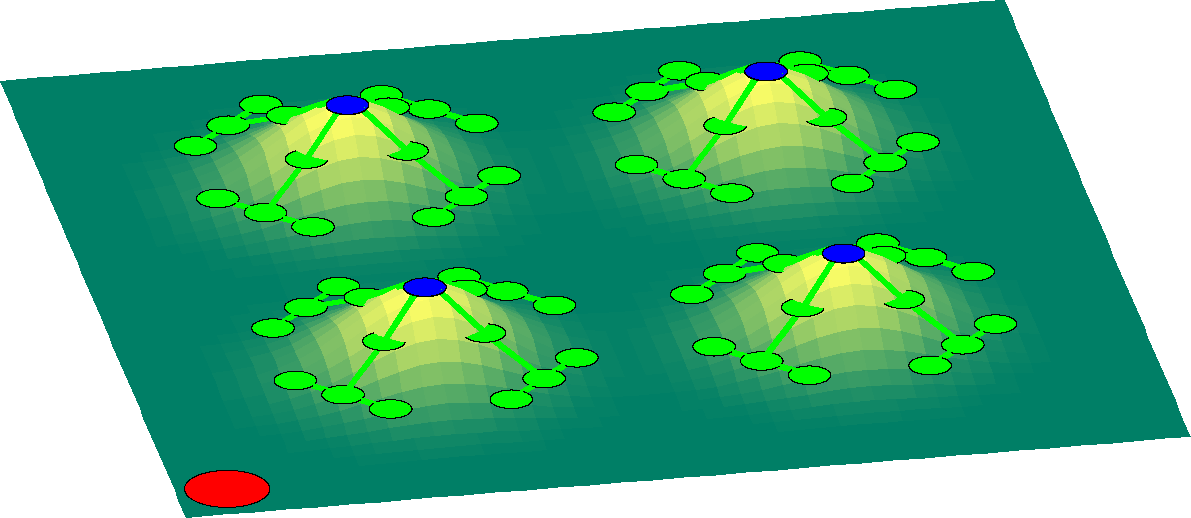}
\caption{Example of the surveillance domain. The red cell in the lower corner is the location of the base station, green cells correspond to surveillance locations, and blue cells are hill tops. Visibility between locations is indicated with a green edge. The base station is tasked with deploying drones to find poachers who may infiltrate at one of the surveillance locations.}
\label{fig:expt2domain}
\end{center}
\end{figure}

Concretely, we consider a $26 \times 26$ cell grid world, which represents the wildlife reserve, and the base station is situated at a fixed location in one corner. We assume that there are 68 target locations of interest, being areas with a particularly high concentration of wildlife. These areas are arranged around four `hills', the tops of which provide better vantage points. Figure~\ref{fig:expt2domain} depicts this setting.

At each episode, the base station deploys the drone to one of the 68 locations. The interpretation of this in BPR is that these 68 target locations each correspond to a different poacher type, or task. 
 For each type, we assume that there is a pre-learnt policy for reaching and surveying that area while dealing with local wind perturbations and avoiding obstacles such as trees. 

The observation signal that the base station receives after each drone deployment  is noise-corrupted information related to the success in identifying an intruder  at the target location  or somewhere nearby (in a diagonally adjacent cell). One exception is when surveying the hill centres which, by corresponding to a high vantage point, provide a weak signal stating that the intruder is in the larger area  around the hill. For a distance $d$ between the region surveyed and the region occupied by the poachers, the signal $R$ received by the agent is
\begin{eqnarray*}
 R & \longleftarrow & \begin{cases} 200-30d+\psi & \mbox{if agent surveys a hilltop and } d \leq 15  \\
				    200-20d+\psi & \mbox{if agent surveys any another location and } d \leq 3 \\
				    \psi & \mbox{otherwise,} \end{cases}
\end{eqnarray*}
where $\psi \sim N(10, 20)$ is Gaussian noise. A higher signal indicates more confidence in having observed a poacher in the region surrounding the target surveillance point.

\begin{figure}[!h]
\begin{center}
\includegraphics[width=\textwidth]{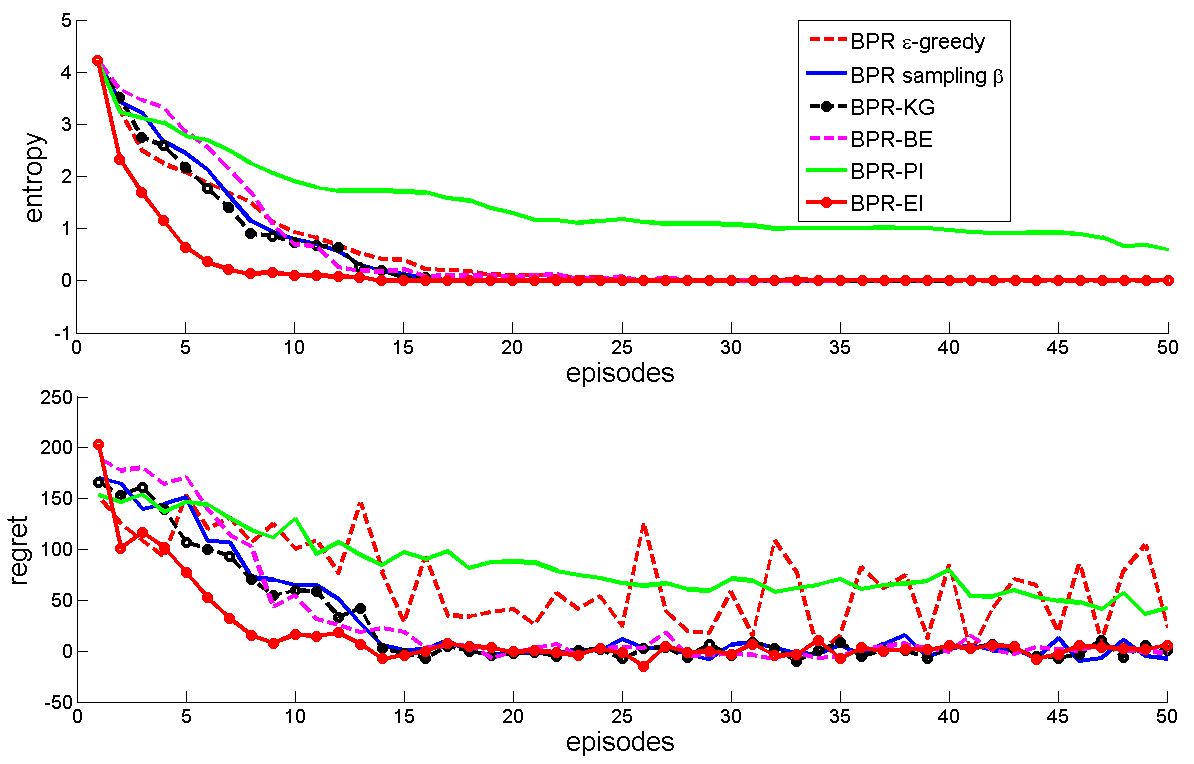}
\caption{Comparison of the six policy selection heuristics on the 68-task surveillance domain, averaged over 10 random tasks. (a) The entropy of the belief  after each episode. (b) The regret after each episode. Error bars are omitted for clarity, but standard deviations of the regret are shown in Table \ref{tab:bprstd}.}
\label{fig:bprvariants}
\end{center}
\end{figure}

Figure~\ref{fig:bprvariants} presents a comparison between six variants of the BPR algorithm. Four use the exploration heuristics proposed in Section \ref{sec:selection}, namely BPR-KG, BPR-BE, BPR-PI, BPR-EI, in addition to \emph{sampling the belief} $\beta$, and \emph{$\epsilon$-greedy} selection with $\epsilon = 0.3$. These six variants were run on the domain 
and averaged over 10 random tasks, with standard deviations of the regret shown in Table \ref{tab:bprstd}.

\begin{table}[!h]
\begin{center}
\begin{tabular}{|c|cccccc|}
\hline
episode & $\epsilon$-greedy & sampling & BPR-KG & BPR-BE & BPR-PI & BPR-EI\\
\hline
5 & 72.193 & 103.07 & 76.577 & 96.529 & 15.695 & 27.801\\
10 & 97.999 & 86.469 & 75.268 & 91.288 & 62.4 & 33.112\\
20 & 83.21 & 18.834 & 7.1152 & 17.74 & 72.173 & 16.011\\
50 & 86.172 & 10.897 & 18.489 & 13.813 & 101.69 & 11.142\\
\hline
\end{tabular}
\caption{Standard deviations of the regret for the six BPR variants shown in Figure~\ref{fig:bprvariants}, after episodes 5, 10, 20 and 50. \label{tab:bprstd}}
\end{center}
\end{table}

Note in Figure~\ref{fig:bprvariants}(a) that BPR-BE, BPR-KG, and BPR with \emph{sampling the belief} all converge in about 15 episodes, which is approximately a quarter the number that would be required by a brute force strategy which involved testing every policy in turn. Both BPR-PI and BPR with \emph{$\epsilon$-greedy} selection fail to converge within the allotted 50 episodes. BPR-EI shows the most rapid convergence.



We now compare the performance of BPR to other approaches from the literature. We choose two frameworks, multi-armed bandits for which we use UCB1 \citep{ucb}, and Bayesian optimisation where we use GP-UCB \citep{srinivas2009gaussian}. We note upfront that although these frameworks share many elements with our own framework in terms of the problems they solve, the assumptions they place on the problem space are different, and thus so is the information they use. 

\begin{figure}[h]
\begin{center}
\includegraphics[width=0.9\textwidth]{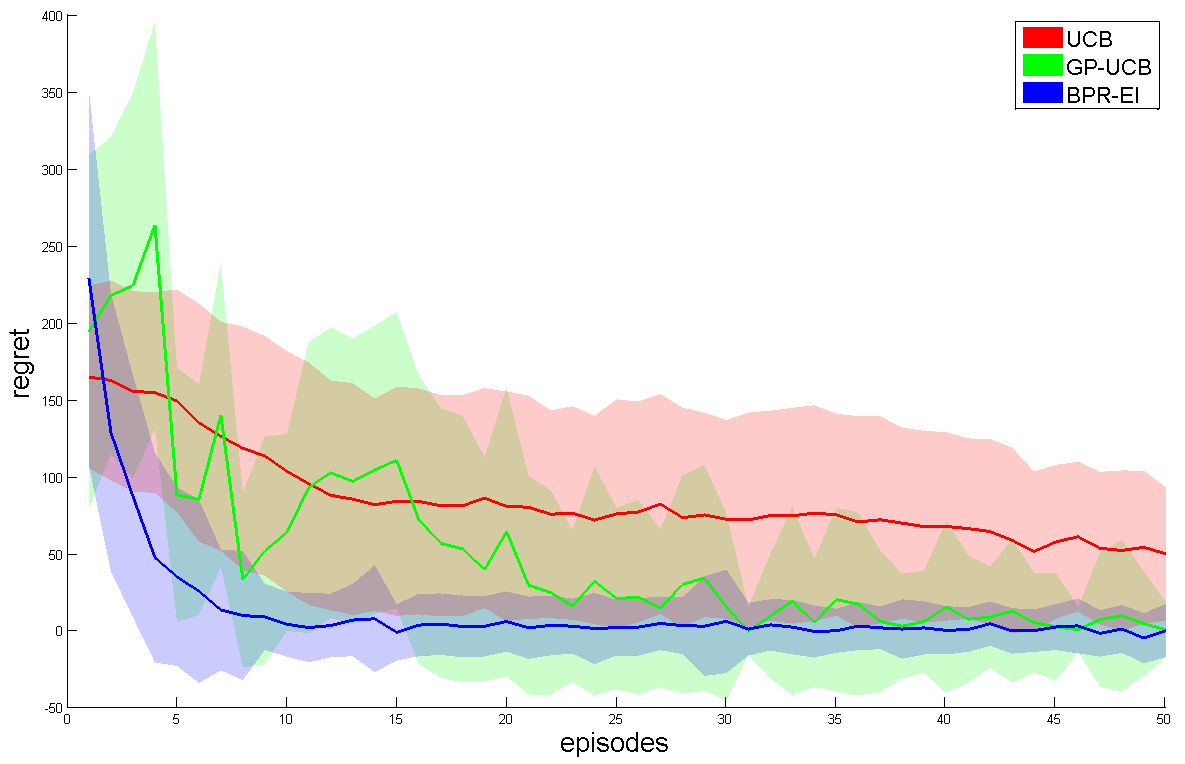}
\caption{Comparison of the episodic regret with time, averaged over 50 random tasks, of BPR-EI, a multi-arm bandits approach (UCB1), and a Bayesian optimisation approach (GP-UCB) on the 68 task surveillance domain. Shaded regions represent one standard deviation.}
\label{fig:bprvsother}
\end{center}
\end{figure}

The results of comparing performance of these approaches are presented in Figure~\ref{fig:bprvsother}  on the surveillance domain, averaged over 50 tasks. We use BPR-EI in this experiment as it was the best performing BPR variant as seen in Figure~\ref{fig:bprvariants}.

For UCB1, we treat each existing policy in the library as a different arm of the bandit. `Pulling' an arm corresponds to executing that policy, and the appropriate reward is obtained. We additionally provide UCB1 with a prior in the form of expected performance of each policy given the task distribution $G_0(\tau)$, which we assumed to be uniform in this case. This alleviates UCB1 from having to test each arm first on the new task (which would require 68 episodes) before it can begin the decision making process. It is still slower to converge than BPR, as information from each episode only allows UCB1 to update the performance estimate of a single policy, whereas BPR can make global updates over the policy space.

On the other hand, an optimisation-based approach such as GP-UCB is better suited to this problem, as it operates with the same requirement as BPR of maintaining low sample complexity. This algorithm treats the set of policies as an input space, and is required to select the point in this space which achieves the best performance on the current task. However, unlike BPR, this approach requires a metric in policy space. This information is not known in this problem, but we approximate this from performance in the training tasks. As a result of this approximation, sampling a single point in GP-UCB (corresponding to executing a policy) again only provides information about a local neighbourhood in policy space, whereas selecting the same action would allow BPR to update beliefs over the entire task space.

Further discussion of the differences between BPR and both bandits and optimisation approaches is provided in Sections \ref{sec:bandits} and \ref{sec:bayesopt} respectively.

\begin{figure}[!h]
\begin{center}
\includegraphics[width=0.9\textwidth]{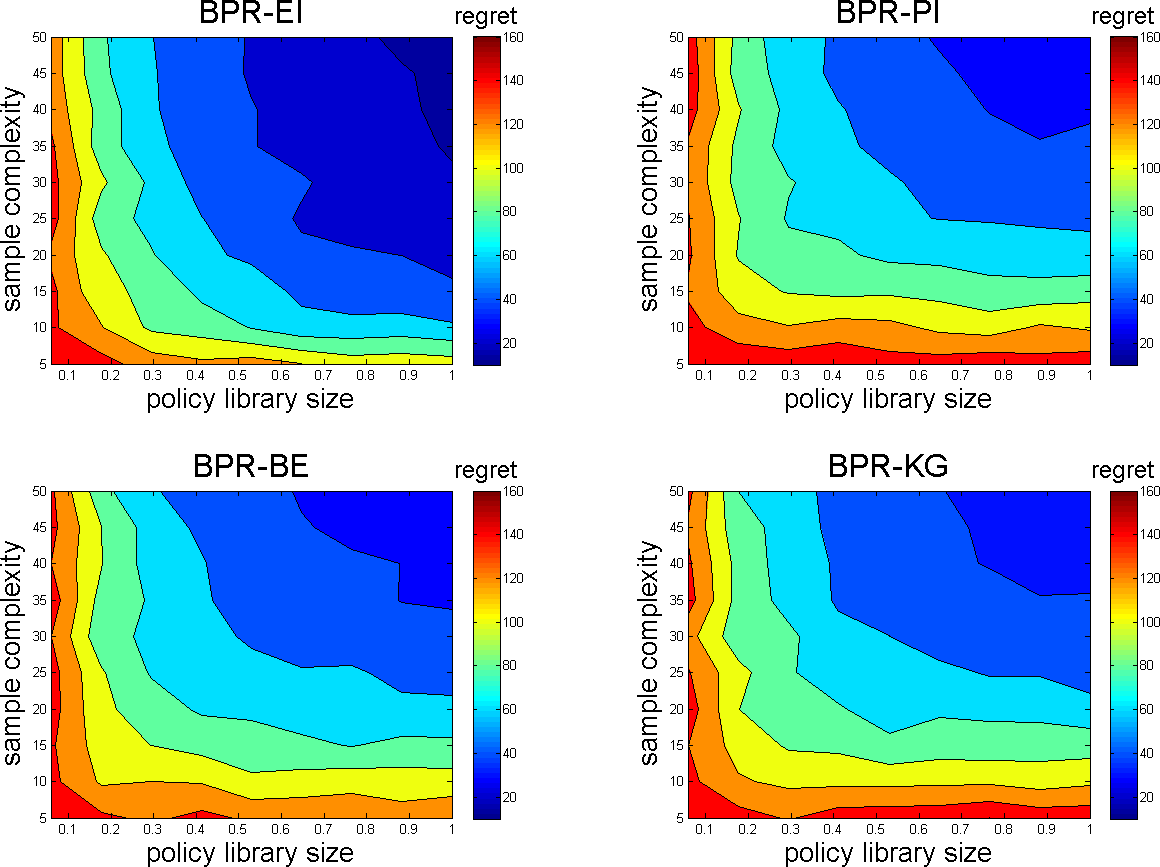}
\caption{Average episodic regret for running BPR-EI, BPR-PI, BPR-BE, and BPR-KG on the 68 task surveillance domain, with different library sizes (as a proportion of the full task space size) and number of episodes (sample complexity), averaged over 200 random tasks.}
\label{fig:librarysize}
\end{center}
\end{figure}

Finally, we explore the trade-off between library size and sample complexity with respect to the regret of BPR-EI, BPR-PI, BPR-BE, and BPR-KG. This is shown in Figure~\ref{fig:librarysize} where, for each method, the horizontal axis shows the ratio of the library size to the full task space size, the vertical axis shows the number of episodes allowed for each new instance, and regret is represented by colour. For each combination  of a library size and a sample complexity, we average the regret results over 200 trials. In each of these trials, a random subset of the full task space is used as the offline policy library and the online task is drawn from the full task space. That is, tasks  in the online phase include both previously-solved and new tasks. 

As can be seen from the figure, regret can be decreased by either increasing  the library size or the time allocated (in terms of number of episodes) to complete the new task. Usually, the task specification dictates the maximum allowed number of episodes, and hence, this suggests a suitable library size to be acquired in the offline phase to attain a specific regret rate. This figure also confirms the previous findings that BPR-EI is able to exceed the other variants in terms of performance.


\section{Discussion and Related Work}

\subsection{Transfer Learning}

The optimal selection from a set of provided policies for a new task is in essence a transfer learning problem~(see the detailed review by~\cite{transfer}). Specifically, Bayesian Policy Reuse aims to select a policy in a library $\Pi$ for transferring to a new, initially unknown, instance. The criterion for this choice is that it is the best policy for the type most similar to the type of the new instance. One transfer approach that considers the similarity between source and target tasks is by \cite{lazaric2008}, where generated $(s,a,r,s')$ samples from the target task are used to estimate similarity to source tasks, which is measured by the average probability of the generated transitions happening under the source task. Then,  samples from the more similar source tasks are used to seed the learning of the target task, while less similar tasks are avoided, escaping negative transfer. More recently, ~\cite{brunskill} consider using the $(s,a,r,s')$ similarity to compute confidence intervals of where, in a 
collection of MDP classes, a new instance best fits. The classes are acquired from experience by clumping together MDPs that do not differ in their transition dynamics or rewards more than a certain level. Once the class is determined, the previous knowledge of that class, in form of dynamics are rewards, is borrowed to inform the process of planning. Bayesian Policy Reuse does not assume learning is feasible, but relies on transferring a useful policy immediately. Also, we use a Bayesian measure of task similarity which allows exploiting prior knowledge of the task space, quickly incorporating observed signals for a faster response, and also, by maintaining beliefs, keeping open the possibility of new MDPs that do not cleanly fit in any of the discovered classes.


\subsection{Correlated and Contextual Bandits}
\label{sec:bandits}


Using a one-off signal per episode relates BPR to Correlated Bandits. In this setting, the decision-making agent is required to pull an arm 
 from a collection of arms, and use its return to update estimates of the arm values of not only the arm that was pulled as in traditional bandits, but of a larger subset of all the arms. In our problem setting, the arms correspond to policies, and the new task instance corresponds to the new bandit `machine' that generates utilities per arm pull (policy execution).

In the Correlated Bandits literature, the form of correlation between the arms is known to the agent. Usually, this happens to be the functional form of the reward curve. The agent's task is then to identify the parameters of that curve, so that the hypothesis of the best arm moves in the reward curve's parameter space. For example, in Response Surface Bandits~\citep{ginebra1995response}, there is a known prior over the parameters of the reward curve and the metric on the policy space is known.  More recently, \cite{mersereau2009} present a greedy policy which takes advantage of the correlation between the arms in their reward functions, assuming a linear form with one parameter, with a known prior. In our work, we approach a space of tasks from a sampling point of view, where an agent experiences sample tasks and uses these to build the models of the domain. Thus we do not assume any functional form for the response surface, and we do not require the metric on the policy space to be known.

In our framework, we only assert assumptions on the continuity and smoothness of the surface. We treat the known types as a set of learnt bandit machines with known behaviour for each of the different arms. These behaviours define local `kernels' on the response surface, which we then approximate by a sparse kernel machine. We track a hypothesis of the best arm using that space. This is to some extent similar to the Gaussian process framework, but in our case  the lack of a metric on the policy space prevents the definition of the covariance functions needed there. This point is elaborated further in Section \ref{sec:bayesopt}.

In another thread, Dependent Bandits~\citep{depndentbandits} assume that the arms in a multi-armed bandit can be clustered into different groups, such that the members of each have correlated reward distribution parameters. Then, each cluster is represented with one representative arm, and the algorithm proceeds in two steps: a cluster is first chosen by a variant of UCB1~\citep{ucb} applied to the set of representative arms, and then the same method is used again to choose between the arms of the chosen cluster. We assume in our work that the set of previously-solved tasks span and represent the space well, but we do not dwell on how this set of tasks can be selected. Clustering is one good candidate for that, and one particular example of identifying the important types in a task space can be seen in the work of \cite{jmlr}.

In Contextual Bandits~\citep{contextual,epoch_greedy}, the agent is able to observe side information (or \emph{context} labels) that are related to the nature of the bandit machine,  and the question becomes one of selecting the best arm for each possible context. Mapping this setting to our problem, a context represents the type, whereas the arms represent the policies. The difference is that in our case the context information (the type label) is latent, and the space of types is not fully known, meaning that the construction of a bounded set of hypotheses of policy correlation under types is not possible. In addition, our setting has that the response of the arms to contexts is only captured through limited offline sampling, but the agent is able to engage with the same context for multiple rounds.

Another related treatment is that of latent bandits~\citep{latent_bandits} where, in the single-cluster arrival case, the experienced bandit machine is drawn from a single cluster with known reward distributions, and in the agnostic case the instances are drawn from many unknown clusters with unknown reward distributions. Our setting fits in between these two extremes, as the instances are drawn from a single, but unknown, cluster with an unknown reward distribution.

\subsection{Relation to Bayesian Approaches}

\subsubsection{Bayesian Optimisation}
\label{sec:bayesopt}
If the problem of Bayesian Policy Reuse is treated as an instance of Bayesian optimisation, we consider the objective
\begin{equation}
\pi^*=\arg \max_{\pi \in \Pi} \mathrm{E}[U|x^*,\pi],\label{eqn:bayesopt}
\end{equation}
where $x^* \in \mathcal{X}$ is the unknown process 
with which the agent is interacting, and $\mathrm{E}[U|x^*, \pi]$ is the expected performance when playing $\pi$ on $x^*$. This optimisation involves a selection from a discrete set of alternative policies ($\pi \in \Pi$), corresponding to sampling from the performance function at a discrete set of locations. However, sampling from this function is expensive (corresponding  to executing a policy for an episode), and as a result the performance function must be optimised in as few samples as possible. 

A Bayesian optimisation solution requires the target function to be modelled as a Gaussian Process (GP). There are two issues here:
\begin{enumerate}
 \item Observations in BPR need not be the performance itself (see Section \ref{sec:signals}), while 
the GP model is appropriate only where these two are the same.
 \item BPR does not assume knowledge of the metric in policy space. This is however required for Bayesian optimisation, so as to define a kernel function for the Gaussian process. An exception is in the case where policies all belong to a parametrised family of behaviours, placing the metric in parameter space as a proxy for policy space. 
\end{enumerate}

Still, we assume smoothness and continuity of the response surface for similar tasks and policies, which is also a standard assumption in Gaussian process regression. Bayesian Policy Reuse uses a belief that tracks the most similar previously-solved type, and then reuses the best policy for that type. This belief can be understood as the mixing coefficient in a mixture model that represents the response surface. 

\begin{figure}[h!]
\centering
\hspace{0.5cm}
\includegraphics[width=0.9\textwidth]{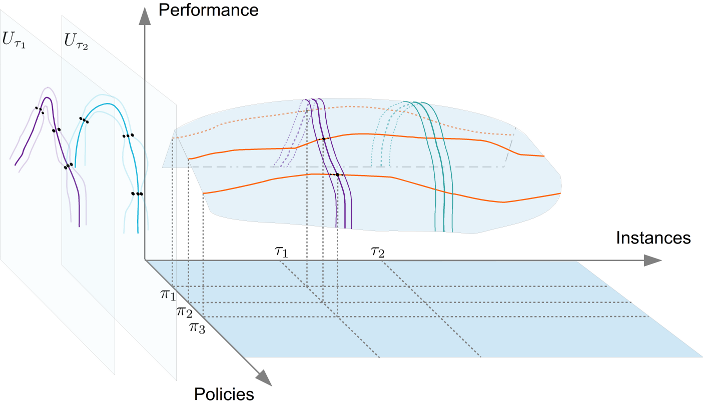} 
\caption{An example 2D response surface. The `bands' on the curve show two types, and the lines that run through the curve from left to right are policy performances for all types. The agent only has access to the intersection of types' bands with policy curves (the black dots). Shown on the left are the performance curves of the two types $\tau_1$ and $\tau_2$ under all policies. These are represented as Gaussian processes in the Policies-Performance plane. Note that Fig. \ref{fig:bpr} is a projection of this response surface. \label{surface}}
\end{figure}

To see this, consider Figure~\ref{surface} which shows an example 2D response surface. Each type is represented by a `band' on that surface; a set of curves only precisely known (in terms of means and variances) at their intersections with a small collection of known policies. Projecting these intersections of some type into performance space results in a probabilistic description of the performance of the different policies on that type (the Gaussian processes in the figure), the kind of function that we are trying to optimise in Equation~\ref{eqn:bayesopt}. Each of these projections would be a component of the mixture model that represents the response surface, and would be the type's contribution to it. 

Any new task instance corresponds to an \emph{unknown} curve on the surface, and correspondingly to a probabilistic model in performance space. Given that the only knowledge possessed by the agent from the surface are these Gaussian processes for each known type, Bayesian Policy Reuse implicitly assumes that they act as a basis that span the space of possible curves, so that the performance under any new task can be  represented as a weighted average of the responses of the previously-solved types\footnote{Note that this will create a bias in the agent's estimated model of the type space toward the types that have been seen more often before. We assume that the environment is benign and that the offline phase is long enough to experience the necessary types.}. To this extent, the performance for the new task instance is approximately identified by a vector of weights, which in our treatment of BPR we refer to as the type belief. Thus, the BPR algorithm is one that fits a probabilistic model to an unknown 
performance curve (Equation~\ref{eqn:bayesopt}) through sampling and weight adjusting in an approximate mixture of Gaussian processes.

\subsubsection{Bayesian Reinforcement Learning}
Bayesian Reinforcement Learning (BRL) is a paradigm of Reinforcement Learning that handles the uncertainty in an unknown MDP in a Bayesian manner by maintaining a probability distribution over the space of possible MDPs, and updating that distribution using the observations generated from the MDP as the interaction continues~\citep{bayesrl}. In work by \cite{mtrl}, the problem of Multi-task Reinforcement Learning of a possibly-infinite stream of MDPs is handled in a Bayesian framework. The authors model the  MDP generative process using a hierarchical infinite mixture model, in which any MDP is assumed to be generated from one of a set of initially-unknown classes, and a hyper-prior controls the distribution of the classes.

Bayesian Policy Reuse can be regarded as an special instance of Bayesian Multi-task Reinforcement Learning with the following construction. Assume a Markov Decision Process that has a chain of $K$ identical states (representing the trials) and a collection of viable actions that connect each state to the next in the chain. The set of actions is given by $\Pi$, the policy library. The processes are parametrised with their type label $\tau$. For each decision step, the agent takes an action (a policy $\pi\in \Pi$) and the process returns with a performance signal, $U_\tau^\pi$. The task of the agent is to infer the best `policy' for this process (a permutation of $K$ policies from $\Pi$; $\pi^0,\ldots,\pi^{K-1}$) that achieves the fastest convergence of values $U$, and thus low convergence time and low regret. The performance/observation models act as the Bayesian prior over rewards required in Bayesian reinforcement learning.

\subsubsection{Other Bayesian Approaches}
\cite{ghavamzadeh07} introduce a Bayesian treatment to the Policy Gradient method in reinforcement learning. The gradient of some parametrised policy space is modelled as a Gaussian process, and paths sampled from the MDP (completed episodes) are used to compute the posteriors  and to optimise the policy by moving in the direction of the performance gradient. The use of Gaussian processes in policy space is similar to the interpretation of our approach, but their use is to model the gradient rather than the performance itself.

When no gradient information is available to guide the search, \cite{wingate2011} propose to use MCMC to search in the space of policies which is endowed with a prior. Various kinds of hierarchical priors that can be used to bias the search are discussed. In our work, we choose the policies using exploration heuristics based on offline-acquired performance profiles rather than using kernels and policy priors. Furthermore, we have access only to a small set of policies to search through in order to optimise the time of response.

\subsection{Storage Complexity}

As described in Section \ref{sec:signals}, the use of different signals entail different observation models and hence different storage complexities. Assume that $|S|$ is the size of the state space, $|A|$ is the size of the action space, $|\Pi|$ is the size of the policy library, $N$ is the number of previously-solved types, $|R|$ is the size of the reward space, $T$ is the duration of an episode, and $B$ is the number of bits needed to store one probability value. For the performance signal, the storage complexity of the observation model is upper bounded by $\mathcal{SC}_{U}=|\Pi|\,N\,|R|\,B$ for the average reward case, and $\mathcal{SC}_{U,\gamma}=|\Pi|\,N\,\frac{1-\gamma^T}{1-\gamma}\,|R|\,B$ for the discounted reward case. For the state-action-state signals, we have $\mathcal{SC}_{s'}=|\Pi|\,N\,|R|\,|S|\,|A|\,B$, and for the immediate reward signal we have $\mathcal{SC}_{r}=|\Pi|\,N\,|S|^2\,|A|\,B$. In applications where $|R|>|S|$ we obtain the ordering $\mathcal{SC}_{U}<\mathcal{SC}_{r}<\mathcal{SC}_{
s'}$.



\section{Conclusion}

In this paper we address the policy reuse problem, which involves responding to an unknown task instance by selecting between a number of policies available to the agent so as to minimise regret, with respect to the best policy in the set, within a short number of episodes. This problem is motivated by many application domains where tasks have short durations  such as human interaction and personalisation or monitoring tasks. 

We introduce Bayesian Policy Reuse, a Bayesian framework for solving this problem. The algorithm tracks a probability distribution (belief) over a set of known tasks capturing their similarity to the new instance that the agent is solving. The belief is updated with the aid of side information (signals) available to the agent: observation signals acquired online for the new instance, and signal models acquired offline for each policy. To balance the trade-off between exploration and exploitation, several mechanisms for selecting policies from the belief (exploration heuristics) are also described, giving rise to different variants of the core algorithm.

This approach is empirically evaluated in three simulated domains where we compare the different variants of BPR, and contrast performance with related approaches. In particular, we compare the performance of BPR with a multi-armed bandit algorithm (UCB1) and a Bayesian optimisation method (GP-UCB). We also show the effect of using different kinds of observation signals on the convergence of the belief, and we illustrate the trade-off between library size and sample complexity required to achieve a required level of performance in a task.

The problem of policy reuse as defined in this paper has many connections with related settings from the literature, especially in the multi-armed bandit research. However, it also has certain features that does not allow it to be reduced exactly to any one of them. The contributed  problem definition and the proposed Bayesian approach are first steps toward a practical solution that can be applied to real world scenarios where traditional learning approaches are not feasible.



\begin{acknowledgements}
This work has taken place in the Robust Autonomy and Decisions group within the School of Informatics, University of Edinburgh. This research has benefitted from support by the UK Engineering and Physical Sciences Research Council (grant number EP/H012338/1) and the European Commission (TOMSY and SmartSociety grants).
\end{acknowledgements}

\bibliographystyle{plainnat}
\bibliography{ref}

\end{document}